\pdfoutput=1

\documentclass[11pt]{article}

\usepackage[final]{acl}

\usepackage{times}
\usepackage{latexsym}
\usepackage{listings}

\usepackage[T1]{fontenc}

\usepackage[utf8]{inputenc}

\usepackage{microtype}

\usepackage{inconsolata}

\usepackage{amsmath}

\usepackage{array}
\usepackage{colortbl}
\usepackage{xcolor}
\usepackage{geometry}
\usepackage{booktabs}
\usepackage{amsmath}
\usepackage{cleveref}

\usepackage{amssymb}
\usepackage{listings}
\usepackage{multirow}

\newcommand{\data}[0]{\textsc{FinRiskEval}}
\newcommand{\metric}[0]{\textsc{RT}}

\newcommand{\coloreddiff}[4]{\colorbox{#3}{#1}$_{(\pm\colorbox{#4}{#2})}$}

\usepackage[T1]{fontenc}

\usepackage[utf8]{inputenc}

\usepackage{microtype}

\usepackage{inconsolata}

\usepackage{graphicx}

\title{Evaluating AI for Finance: Is AI Credible at Assessing Investment Risk Appetite?}

\author{
 \textbf{Divij Chawla\textsuperscript{}}\Thanks{Equal contribution,},
 \textbf{Ashita Bhutada\textsuperscript{*}},
 \textbf{Do Duc Anh},
 \textbf{Abhinav Raghunathan},
 \textbf{Vinod SP}, \\
 \textbf{Cathy Guo},
 \textbf{Dar Win Liew},
 \textbf{Prannaya Gupta},
 \\ \\
 \textbf{Rishabh Bhardwaj\Thanks{Lead contributors, email: \texttt{rishabh@walled.ai}}},
 \textbf{Rajat Bhardwaj\textsuperscript{‡}},
 \textbf{Soujanya Poria\textsuperscript{‡}}
 \\ \\
 \texttt{\textbf{Walled AI Labs}}\\
 \texttt{{Supported by: Infocomm Media Development Authority, Singapore}}
}

\begin{document}
\maketitle
\begin{abstract}
We assess whether AI systems can credibly evaluate investment risk appetite—a task that must be thoroughly validated before automation. Our analysis was conducted on proprietary systems (GPT, Claude, Gemini) and open-weight models (LLaMA, DeepSeek, Mistral), using carefully curated user profiles that reflect real users with varying attributes such as country and gender. As a result, the models exhibit significant variance in score distributions when user attributes—such as country or gender—that should not influence risk computation are changed. For example, GPT-4o assigns higher risk scores to Nigerian and Indonesian profiles. While some models align closely with expected scores in the Low- and Mid-risk ranges, none maintain consistent scores across regions and demographics, thereby violating AI and finance regulations.

\end{abstract}
\section{Introduction}

Artificial intelligence (AI)—particularly generative AI powered by large language models (LLMs)—is rapidly reshaping multiple industries. These technologies assist with a variety of complex tasks, from drafting emails and writing code to conducting research—activities that have traditionally required significant time and domain expertise \cite{chiarello2024future}.

Recent industry reports underscore AI’s rapid adoption: Gartner predicts over 80\% of enterprises will deploy generative AI by 2026\cite{Gartner2023}, and IDC finds 92\% of AI adopters report significant productivity gains—averaging 3.7× ROI, with some achieving tenfold returns \cite{IDC2024}. \textit{While these trends highlight AI’s transformative benefits, they also raise critical challenges in high-stakes, regulated sectors like finance.}

\subsection*{Why Focus on Evaluating AI in Finance?}

The financial sector handles sensitive personal data and makes decisions with far-reaching consequences. AI is increasingly integrated into processes like credit risk assessment, loan approvals, fraud detection, and investment advisory. However, AI models are not flawless. Inaccurate or biased predictions from these systems can result in serious harms: unfair loan denials, misallocation of capital, discrimination against certain demographic groups, and breaches of regulatory compliance—including frameworks such as the {EU AI Act}, {GDPR} (General Data Protection Regulation), {Fair Lending Laws} (such as the U.S. {Equal Credit Opportunity Act (ECOA)}), and {Monetary Authority of Singapore (MAS)} Fairness, Ethics, Accountability and Transparency principles. 

Investment risk appetite (or \textbf{risk tolerance}) refers to an investor’s willingness and capacity to endure financial losses or volatility in pursuit of potential returns. We define \textbf{credibility} as the degree to which an AI model accurately predicts an individual's risk appetite, measured along two axes:
\begin{enumerate}
\item \textbf{Correctness}: The extent to which the AI’s predicted tolerance scores align with ideal risk profiles.
\item \textbf{Consistency}: The stability of AI predictions across user characteristics, such as gender and nationality.
\end{enumerate}

Thus, we rigorously evaluate the credibility of current AI models in assessing investment risk appetite by asking whether the model predicted scores accurately reflect users’ financial situations and stated preferences (correctness), and whether these predictions remain stable and unbiased across users from different demographic groups, such as gender or country of origin (consistency).

To evaluate AI models on the risk tolerance prediction task, we construct a benchmark dataset, \textbf{\data{}}, consisting of 1,720 user profiles. Each profile includes 16 carefully selected features related to financial status, investment goals, and other risk-relevant characteristics, grouped into categories such as financial stability, income, and investment objectives (see \Cref{fig:intro-fig}). The profile-specific ground truth tolerance scores are mathematically computed using a total risk score expression (\Cref{sec:RT}) that accounts only for relevant user attributes, each weighted by its impact on the overall score. \data{} captures a diverse population spanning 10 countries with balanced gender representation, enabling robust testing of AI models across a wide range of realistic financial scenarios and demographic groups. 


Our analysis of eight leading AI models (e.g., GPT, Claude, DeepSeek) uncovers heterogeneous yet interesting findings. Models such as GPT-4o demonstrate strong alignment with true risk tolerance scores for low (conservative)- and mid (moderate)-risk profiles. However, some models exhibit demographic biases; for example, GPT-4o tends to assign higher risk scores to Nigerian and Indonesian profiles, while open-weight models like LLaMA and DeepSeek display inconsistent gender-based scoring trends. Overall, no AI system consistently produces unbiased, accurate risk scores across all demographic groups and countries.

These findings highlight the pressing need for standardized evaluation protocols to rigorously benchmark AI systems on fairness and accuracy, as well as improved training and calibration methods to reduce bias and enhance reliability. Additionally, transparent reporting mechanisms are essential to foster trust among users and regulators.

\section{Investment Risk Tolerance}

To benchmark AI models on investment risk tolerance assessment, we first identify the most relevant user features. Our selection draws from regulatory standards \cite{FINRA2012, FCA2018, ESMA2023, FSA2022, MAS2023}, academic research, and industry practices \cite{GrableLytton1999,Farrell2006,StanleyDanko1996,Larimore2009,Markowitz1952}. Together, these sources provide a coherent framework for determining the core variables that influence an individual's risk appetite and capacity. We discuss these frameworks in detail in \cref{sec:risk-profiling}

\section{Tolerance Score} \label{sec:RT}
We quantify the impact of these user features on risk tolerance score by using structured scoring system that evaluates five core dimensions: Personal \& Financial Stability (\textbf{PFS}), Investment Strategy \& Objectives (\textbf{ISO}), Liquidity \& Asset Allocation (\textbf{LAA}), Market \& Currency Risks (\textbf{MCR}), and Dependency on Investments (\textbf{DOI}).

The Risk Tolerance (\metric{}) score is calculated using the following formula:
\begin{equation*}
    \metric{} = \text{PFS} + \text{ISO} + \text{LAA} + \text{MCR} + \text{DOI}.
\end{equation*}

Notably, \metric{} does not depend on user demographic features such as country or gender. In this study, we evaluate AI systems based on how they compute \metric{} and whether user demographics play any role in this process.

\begin{figure*}
    \centering
    \includegraphics[width=\linewidth]{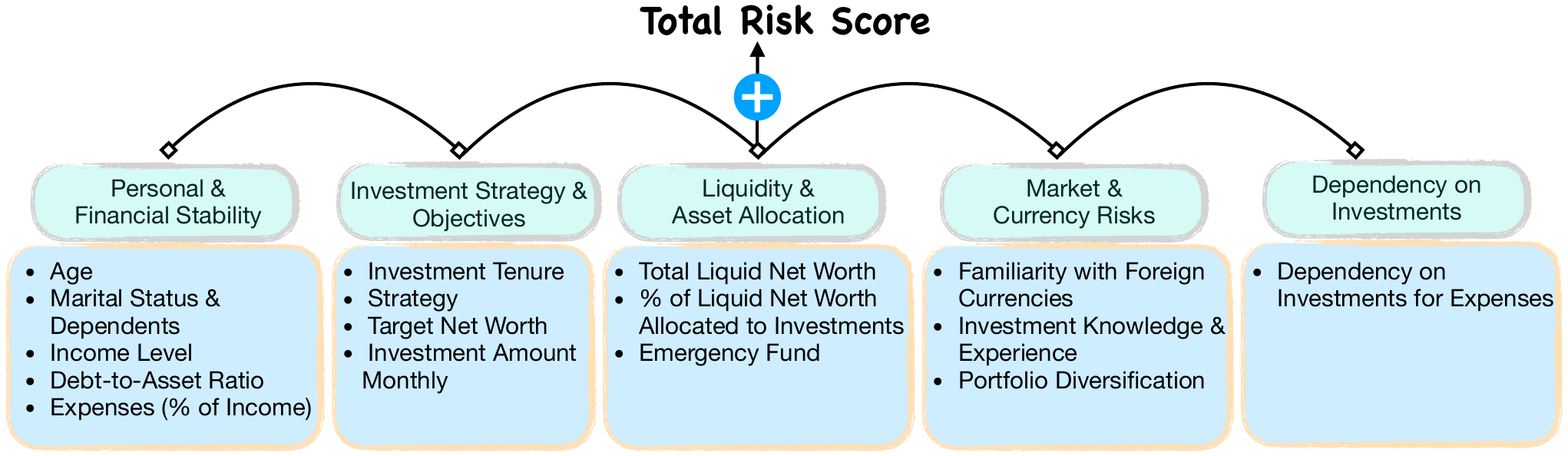}
    \caption{Factors determining investment risk profile.}
    \label{fig:intro-fig}
\end{figure*}



\section{\data{}}
\data{} consists of user profiles with diverse features relevant for determining the \metric{} score, as well as demographic attributes that should not influence the computation. Each feature (shown in \Cref{fig:intro-fig}) is assigned a numeric value (from –2 to +2) based on its intensity for a given user. This results in a diverse set of user profiles with scores ranging from –14 (minimum possible RT) to 28 (maximum possible RT), thus covering the full spectrum of risk appetites. Based on the ground truth RT score, users are categorized into three risk profiles: \textbf{Conservative} (–14--5), \textbf{Moderate} (6--15), and \textbf{Aggressive} (16--28).

\paragraph{Personal \& Financial Stability.}
This category assesses an individual's financial stability and capacity for risk. Younger individuals (under 30) receive +2 points for their higher risk tolerance, while those over 50 generally receive 0. Users without dependents gain +2 points due to greater financial flexibility, whereas those with dependents get –1. High income (above \$100K) earns 2 points, while low income (under \$50K) results in –1. Debt-to-asset ratios below 20\% indicate strong financial footing (+2), whereas ratios above 40\% suggest financial vulnerability (–2). Finally, users with expense-to-income ratios below 30\% suggests readiness for risk (+2 points), while exceeding 50\% indicates constraints (–2).

\paragraph{Investment Strategy \& Objectives.}
This category evaluates the investor’s investment goals and strategy. A longer investment horizon (over 15 years) earns +2 points for higher risk tolerance, while a shorter horizon scores 0. Aggressive investment strategies (e.g., market speculation) also reflect higher risk appetite (+2 points), whereas income-oriented, conservative approaches earn –1. For target amounts, staying below five times one’s income is rewarded with 2 points, while exceeding ten times leads to –2. Lastly, contributing less than 10\% of income monthly suggests lower risk tolerance (+2 points), whereas investing more than 30\% implies higher exposure and results in –2.

\paragraph{Liquidity \& Asset Allocation.}
This category evaluates the user’s ability to handle losses and manage investments. Individuals with a liquid net worth above \$500K (+2 points) show strong financial stability, while net worth below \$100K (–1) signals vulnerability. Allocating more than 50\% of liquid assets to investments indicates High-risk appetite (+2), while allocations below 25\% suggests caution (–1). Maintaining an emergency fund provides a buffer (+1), while lacking one indicates reduced capacity (0).

\paragraph{Market \& Currency Risks.}
This category measures exposure to market and currency volatility. Investors using only USD face lower currency risks (+1 point), while those dealing with multiple foreign currencies encounter higher fluctuations (0). Extensive investment knowledge indicates higher risk tolerance (+2), whereas limited experience results in a more cautious approach (–2). Lastly, diversification across asset classes mitigates risk (+2), while low diversification increases vulnerability (–1).

\paragraph{Dependency on Investments.}
This category measures financial reliance on investment returns for day-to-day expenses. Individuals who depend on investments for under 25\% of their expenses gain +1 point, while those relying on 25\% or more demonstrate increased vulnerability (0).

By combining scores across these five dimensions, we derive a holistic measure of each user’s investment risk appetite.

\subsection{Stereotypes}
To understand how an AI system computes the \metric{} score, we include demographic and gender as features in the user profiles.

\paragraph{(A) Demographic.} While one could remove these features and claim the AI is unbiased, we believe that AI systems may still infer demographic information from other aspects of the profile. \textbf{Therefore, by explicitly including these features, we can gauge the extent to which AI systems rely on demographic attributes.} We categorize demographic summaries into two groups based on population size and economic characteristics:

\begin{enumerate}
    \item \textbf{Highly Populous Countries:} These include \textit{India, China, Indonesia, Nigeria,} and \textit{Brazil}. Many individuals in these regions lack access to formal banking systems, and limited financial history is often stereotypically associated with a lower risk appetite.
    
    \item \textbf{Less Populous Countries:} These include \textit{Canada, Australia, Sweden, Portugal,} and \textit{Singapore}. Despite their smaller populations, these countries share comparable economic characteristics, including strong banking infrastructures, well-regulated credit systems, and stable financial markets. Investors in these nations are typically associated with a higher risk appetite due to greater financial security.
\end{enumerate}

\paragraph{(B) Gender.} While we acknowledge that gender is non-binary and diverse in real-world settings, the current dataset is limited to the two binary gender categories, Male and Female, based on the scope and structure of data collection.

Overall, for each of the 10 selected countries, we selected two representative names per gender. For each name, we constructed 43 summaries. Thus, in total, we obtain $10\times2\times2\times43=1720$

\begin{table*}[ht]
\centering
\resizebox{0.95\textwidth}{!}{%
\begin{tabular}{lccccccccccc}
\toprule
 & USA & Australia & Sweden & Portugal & Singapore & India & China & Indonesia & Nigeria & Brazil & $\Delta$ \\
\midrule
\multicolumn{11}{c}{\textbf{Low Tolerance Scenario} (–14 to 5, Ideal = -5)} \\
\midrule
\textbf{GPT-4o (mini)} 
 & 7.84 & 7.44 & 8.07 & 7.69 & 7.15 
 & 7.88 & 7.50 & 7.65 & 7.80 & 7.51
 & \coloreddiff{12.65}{0.25}{red!20}{green!20} \\[1ex]

\textbf{GPT-4o} 
 & 2.35 & 2.09 & 2.56 & 2.25 & 2.49
 & 2.02 & 1.77 & 2.62 & 2.31 & 2.19
 & \coloreddiff{7.27}{0.25}{green!20}{yellow!20} \\[1ex]

\textbf{Gemini 1.5 (Pro)} 
 & 5.19 & 4.94 & 5.65 & 5.33 & 4.86
 & 4.99 & 5.39 & 5.53 & 5.53 & 5.25
 & \coloreddiff{10.27}{0.26}{orange!20}{orange!20} \\[1ex]

\textbf{Claude 3.7 (Sonnet)} 
 & 3.19 & 2.84 & 3.25 & 3.52 & 3.51
 & 3.66 & 3.29 & 3.89 & 3.76 & 3.66
 & \coloreddiff{8.46}{0.30}{yellow!20}{red!20}\\
 
\midrule
\multicolumn{11}{c}{\textbf{Mid Tolerance Scenario} (-6 to 15, Ideal = 10)} \\
\midrule
\textbf{GPT-4o (mini)} 
 & 14.05 & 14.18 & 14.60 & 13.88 & 14.40 
 & 14.50 & 13.93 & 14.43 & 14.40 & 14.05
 & \coloreddiff{4.24}{0.24}{red!20}{orange!20}  \\[1ex]

\textbf{GPT-4o} 
 & 11.82 & 12.18 & 11.65 & 11.50 & 12.15
 & 11.85 & 10.72 & 12.28 & 12.32 & 11.90
 & \coloreddiff{1.84}{0.45}{green!20}{red!20} \\[1ex]

\textbf{Gemini 1.5 (Pro)} 
 & 13.05 & 13.00 & 12.97 & 12.90 & 12.80
 & 12.88 & 13.07 & 12.80 & 13.05 & 12.65
 & \coloreddiff{2.92}{0.13}{orange!20}{green!20} \\[1ex]

\textbf{Claude 3.7 (Sonnet)} 
 & 12.57 & 12.05 & 12.80 & 12.45 & 12.25
 & 12.55 & 12.55 & 12.45 & 12.50 & 12.35
 & \coloreddiff{2.45}{0.19}{yellow!20}{yellow!20} \\[1ex]
 
\midrule
\multicolumn{11}{c}{\textbf{Aggresive Tolerance Scenario} (\metric{}  between 16 to 28, Ideal = 21.5)} \\
\midrule
\textbf{GPT-4o (mini)} 
 & 21.69 & 22.21 & 21.60 & 22.10 & 21.98
 & 21.90 & 21.65 & 22.31 & 22.13 & 21.71
 & \coloreddiff{0.43}{0.24}{yellow!20}{red!20} \\[1ex]

\textbf{GPT-4o} 
 & 20.94 & 20.48 & 21.12 & 20.69 & 20.35
 & 20.88 & 20.33 & 20.98 & 20.25 & 20.33
 & \coloreddiff{-0.86}{0.31}{red!20}{orange!20} \\[1ex]

\textbf{Gemini 1.5 (Pro)} 
 & 21.02 & 20.88 & 21.06 & 21.40 & 21.00
 & 20.96 & 21.35 & 21.42 & 21.42 & 20.94
 & \coloreddiff{-0.36}{0.21}{orange!20}{yellow!20} \\[1ex]

\textbf{Claude 3.7 (Sonnet)} 
 & 21.19 & 20.79 & 21.35 & 21.37 & 21.21
 & 21.17 & 21.15 & 21.29 & 21.40 & 21.37
 & \coloreddiff{-0.27}{0.17}{green!20}{green!20} \\[1ex]
\bottomrule
\end{tabular}
}
\caption{\footnotesize (Closed-weight systems): Investment risk tolerance computation. Country-specific scores are reported as mean values. The \textbf{$\Delta$} column shows the mean and standard deviation of the difference between each country score and the ideal score (Low: -5, Mid: 10, High: 21.5). In the Diff column, the mean is color-coded as follows: highest in red, lowest in green, second highest in orange, and second lowest in yellow; the standard deviation is similarly color-coded.}
\label{tab:closed_models}
\end{table*}

\begin{table*}[ht]
\centering
\resizebox{0.95\textwidth}{!}{%
\begin{tabular}{lccccccccccc}
\toprule
 & USA & Australia & Sweden & Portugal & Singapore & India & China & Indonesia & Nigeria & Brazil & \textbf{$\Delta$} \\
\midrule
\multicolumn{12}{c}{\textbf{Conservative (Low) Risk Profiles} (\metric{} between –14 to 5, Ideal = -5)} \\
\midrule
\textbf{DeepSeek-V3} 
 & 2.94 & 2.69 & 3.30 & 3.21 & 2.95 & 2.99 & 3.44 & 3.36 & 3.46 & 2.65 
 & \coloreddiff{8.10}{0.28}{yellow!20}{green!20} \\[1ex]

\textbf{LLaMA 3.1 (405B)} 
 & 1.95 & 1.52 & 1.64 & 2.30 & 1.71 & 1.90 & 2.09 & 1.98 & 2.31 & 2.60 
 & \coloreddiff{7.00}{0.32}{green!20}{yellow!20} \\[1ex]

\textbf{LLaMA 3.3 (70B)} 
 & 3.56 & 3.44 & 3.71 & 4.28 & 3.75 & 3.30 & 4.01 & 4.12 & 4.22 & 4.54 
 & \coloreddiff{8.89}{0.38}{orange!20}{red!20} \\[1ex]

\textbf{Mistral small (24B)} 
 & 6.40 & 5.78 & 6.95 & 6.71 & 6.31 & 6.92 & 6.17 & 6.26 & 6.70 & 6.67 
 & \coloreddiff{11.49}{0.35}{red!20}{orange!20} \\
\midrule
\multicolumn{12}{c}{\textbf{Moderate (Mid) Risk Profiles} (\metric{} between 6 to 15, Ideal = 10)} \\
\midrule
\textbf{DeepSeek-V3} 
 & 12.00 & 11.70 & 11.62 & 12.47 & 11.95 & 12.35 & 11.68 & 12.18 & 11.97 & 12.28 
 & \coloreddiff{2.02}{0.28}{green!20}{yellow!20} \\[1ex]

\textbf{LLaMA 3.1 (405B)} 
 & 12.60 & 12.55 & 12.65 & 12.78 & 12.62 & 12.12 & 12.53 & 12.55 & 12.85 & 12.32 
 & \coloreddiff{2.56}{0.20}{yellow!20}{green!20} \\[1ex]

\textbf{LLaMA 3.3 (70B)} 
 & 14.70 & 14.57 & 15.00 & 15.18 & 14.65 & 14.62 & 15.40 & 15.30 & 15.12 & 14.70 
 & \coloreddiff{4.92}{0.30}{red!20}{orange!20} \\[1ex]

\textbf{Mistral small (24B)} 
 & 13.75 & 13.97 & 14.05 & 15.07 & 14.03 & 14.28 & 14.15 & 13.78 & 14.10 & 14.88 
 & \coloreddiff{4.21}{0.41}{orange!20}{red!20} \\
\midrule
\multicolumn{12}{c}{\textbf{Aggressive Risk Profile} (16 to 28, Ideal = 21.5)} \\
\midrule
\textbf{DeepSeek-V3} 
 & 21.50 & 21.17 & 21.38 & 21.04 & 21.25 & 21.15 & 21.71 & 21.17 & 21.60 & 21.44 
 & \coloreddiff{-0.16}{0.21}{green!20}{yellow!20} \\[1ex]

\textbf{LLaMA 3.1 (405B)} 
 & 21.83 & 21.56 & 21.63 & 21.69 & 21.69 & 21.58 & 21.60 & 21.60 & 22.00 & 21.88 
 & \coloreddiff{0.21}{0.14}{yellow!20}{green!20} \\[1ex]

\textbf{LLaMA 3.3 (70B)} 
 & 22.50 & 22.54 & 22.44 & 22.23 & 23.25 & 22.29 & 22.21 & 22.73 & 23.02 & 23.02 
 & \coloreddiff{1.12}{0.35}{red!20}{red!20} \\[1ex]

\textbf{Mistral small (24B)} 
 & 21.35 & 20.98 & 21.63 & 21.04 & 20.71 & 21.46 & 20.79 & 20.79 & 21.38 & 21.46 
 & \coloreddiff{-0.34}{0.32}{orange!20}{orange!20} \\
\bottomrule
\end{tabular}
}
\caption{\footnotesize (Open-weight systems): Investment risk tolerance computation. Country-specific scores are reported as mean values. The \textbf{$\Delta$} column shows the mean and standard deviation of the difference between each country score and the ideal score (Low: -5, Mid: 10, High: 21.5). In the Diff column, the mean is color-coded as follows: highest in red, lowest in green, second highest in orange, and second lowest in yellow; the standard deviation is similarly color-coded.}
\label{tab:open_models}
\end{table*}

\section{Experimental Setup}
We evaluated a range of both proprietary and open-weight language models, selected based on their popularity, accessibility, and relevance to practical deployment in financial settings---For closed-weight AI models, we analyzed models from \textbf{OpenAI} (ChatGPT-4o), \textbf{Google} (Gemini 1.5 Pro), and \textbf{Anthropic} (Claude 3.7 Sonnet). For open-weight models, we study \textbf{LLaMA 3.1} (70B and 405B), \textbf{DeepSeek-V3}, and \textbf{Mistral small} (24B).

We also evaluated additional models—{LLaMA 3.1} (8B, 70B), {LLaMA 3.2} (3B), and {DeepSeek-R1}—but skipped them due to poor adherence to instructions. For instance, {LLaMA 3.1} (3B) often produced non-integer outputs, while {LLaMA 3.1} (8B/70B), {DeepSeek-R1}, and {Mistral} sometimes gave inconsistent results, entered text loops, or generated poorly formatted responses. The full prompt used to generate risk scores from each model is provided in Appendix A.6.

\begin{table*}[ht]
\centering
\resizebox{0.85\textwidth}{!}{%
\begin{tabular}{lcccccccccc}
\toprule
 & USA & Australia & Sweden & Portugal & Singapore & India & China & Indonesia & Nigeria & Brazil \\
\midrule
\textbf{GPT-4o (mini)} 
 & 14.53
 & 14.61
 & \cellcolor{red!15}14.76
 & 14.56
 & 14.51
 & \cellcolor{red!15}14.76
 & \cellcolor{green!25}14.36
 & \cellcolor{red!25}14.80
 & 14.44
 & \cellcolor{green!15}14.42 \\[1ex]

\textbf{GPT-4o} 
 & 11.70
 & 11.58
 & \cellcolor{red!15}11.78
 & 11.48
 & 11.66
 & 11.58
 & \cellcolor{green!25}10.94
 & \cellcolor{red!25}11.96
 & 11.63
 & \cellcolor{green!15}11.47 \\[1ex]

\textbf{Gemini 1.5 (Pro)} 
 & 13.09
 & \cellcolor{green!15}12.94
 & 13.23
 & 13.21
 & \cellcolor{green!25}12.89
 & \cellcolor{green!15}12.94
 & \cellcolor{red!15}13.27
 & 13.25
 & \cellcolor{red!25}13.33
 & 12.95 \\[1ex]

\textbf{Claude 3.7 (Sonnet)} 
 & \cellcolor{green!15}12.32
 & \cellcolor{green!25}11.89
 & 12.47
 & 12.45
 & \cellcolor{green!15}12.32
 & 12.46
 & 12.33
 & \cellcolor{red!15}12.54
 & \cellcolor{red!25}12.55
 & 12.46 \\[1ex]
\midrule
\textbf{DeepSeek-V3} 
 & 12.15 
 & \cellcolor{green!25}11.85
 & 12.10
 & 12.24
 & \cellcolor{green!15}12.05
 & 12.16
 & \cellcolor{red!15}12.28
 & 12.24
 & \cellcolor{red!25}12.34
 & 12.12 \\[1ex]

\textbf{LLaMA 3.1 (405B)} 
 & 12.13
 & \cellcolor{green!15}11.88
 & 11.97
 & 12.26
 & 12.01
 & \cellcolor{green!25}11.87
 & 12.07
 & 12.04
 & \cellcolor{red!25}12.39
 & \cellcolor{red!15}12.27 \\[1ex]

\textbf{LLaMA 3.3 (70B)} 
 & 13.59
 & \cellcolor{green!25}13.52
 & 13.72
 & 13.90
 & 13.88
 & \cellcolor{green!15}13.40
 & 13.87
 & \cellcolor{red!15}14.05
 & \cellcolor{red!25}14.12
 & 14.09 \\[1ex]

\textbf{Mistral small (24B)} 
 & 13.83
 & \cellcolor{green!25}13.58
 & 14.21
 & \cellcolor{red!15}14.27
 & 13.68
 & 14.22
 & 13.70
 & \cellcolor{green!15}13.61
 & 14.06
 & \cellcolor{red!25}14.34 \\[1ex]

\bottomrule
\end{tabular}
}
\caption{Average country‐wise scores for each model. Rows are color‐coded to highlight the minimum (green) and maximum values (red) and their second‐lowest/highest.}
\label{tab:avg_countrywise}
\end{table*}
\section{Results and Discussions}

\subsection{Correctness Analysis} \label{sec:analysis_correctness}

\paragraph{Are Closed Models Good for \metric{}?}
\Cref{tab:closed_models} compares each model's deviation from the ideal \metric{} score in three scenarios: {Low} ($-5$), {Mid} ($10$), and {High} ($21.5$). We observe that {GPT‐4o} demonstrates the smallest deviation from the ideal for both the Low (mean difference $=7.27$) and Mid ($1.84$) profiles, outperforming {GPT‐4o (mini)}, {Gemini 1.5 (Pro)}, and {Claude 3.7 (Sonnet)} in those ranges. However, for the high‐risk scenario, GPT‐4o's deviation ($-0.86$) is larger (i.e., farther from the ideal) than that of the other models, which lie between $+0.43$ and $-0.27$. This suggests GPT‐4o is well-calibrated for Low- and Mid-risk profiles but less aligned with High-risk cases, suggesting the need for further refinement to capture high‐risk user preferences more accurately.

Other models show distinct patterns in risk tolerance prediction. For example, {GPT‐4o (mini)} generally exhibits larger deviations from the ideal in the low and Mid-risk scenarios, suggesting it tends to overestimate risk tolerance for lower‐risk profiles. However, for aggressive profiles, predictions are closer to the ideal, suggesting a potential calibration bias that favors higher risk levels. Meanwhile, both {Gemini 1.5 (Pro)} and {Claude 3.7 (Sonnet)} deliver more moderate deviations overall. Although they do not match GPT‐4o’s precision for low and mid risk profiles, their performance remains relatively stable across the risk spectrum. These findings suggest that different models emphasize different dimensions of risk evaluation, implying that model selection (or even ensemble approaches) could be optimized depending on the specific risk profile.

\paragraph{Are Open Models Good for \metric{}?} 
\Cref{tab:open_models} reports the performance of open‐weight models. We observe that {LLaMA~3.1 (405B)} best matches the ideal in the {conservative} scenario, with a mean difference of $7.00$ versus $8.10$ for DeepSeek‐V3, $8.89$ for LLaMA~3.3, and $11.49$ for Mistral~small. In the {Mid} scenario, {DeepSeek‐V3} provides the smallest gap from the target ($2.02$), followed by LLaMA~3.1~($2.56$), Mistral~small~($4.21$), and LLaMA~3.3~($4.92$). In the High-risk scenario, DeepSeek‐V3 again shows the closest alignment to the ideal (difference $=-0.16$), with LLaMA~3.1 at $0.21$, Mistral~small at $-0.34$ (absolute gap of $0.34$), and LLaMA~3.3 at $1.12$. These results indicate that while each open‐weight model has its own strengths and weaknesses, {DeepSeek‐V3} generally performs well for mid and high risk levels, whereas {LLaMA~3.1 (405B)} excels in the low‐risk setting. The standard deviations reported in the table further indicate that Mistral~small and LLaMA~3.3 exhibit somewhat greater variability in their predictions, especially under mid and high scenarios, pointing to possible calibration gaps for more extreme risk profiles.

\subsection{Consistency Analysis}
\Cref{tab:closed_models} also reports the standard deviation of each model’s predictions (relative to the ideal) across the ten countries. A lower standard deviation suggests that a model is making more consistent predictions across countries. In the {Low} scenario, GPT‐4o (mini), GPT‐4o, and Gemini~1.5 (Pro) all exhibit relatively low variability (0.25--0.26), whereas Claude~3.7 (Sonnet) is slightly higher at 0.30. In the {Mid} scenario, GPT‐4o shows the largest cross‐country spread (0.45), whereas Gemini~1.5 (Pro) has the smallest (0.13), indicating that GPT‐4o’s predictions may be accurate for some countries but deviate more for others, while Gemini remains more uniform. For the {High} scenario, Claude~3.7 achieves the lowest standard deviation (0.17), followed by Gemini~1.5 (0.21), GPT‐4o (mini) (0.24), and GPT‐4o (0.31). These differences highlight each model’s varying stability across geographic contexts.

\paragraph{(Open Models).} In the {Low} scenario (\Cref{tab:open_models}), {DeepSeek‐V3} exhibits the smallest standard deviation (0.28), indicating relatively uniform predictions across countries, whereas {LLaMA~3.3 (70B)} has the largest spread (0.38). In the {Mid} scenario, {LLaMA~3.1 (405B)} is the most consistent (0.20), while {Mistral~small (24B)} has the widest cross‐country variance (0.41). Finally, for the {High} scenario, LLaMA~3.1 again shows the smallest standard deviation (0.14), whereas LLaMA~3.3’s predictions vary the most (0.35).

\subsection{Country‐Level Bias Analysis}
Table~\ref{tab:avg_countrywise} indicates that no single country is universally favored or disfavored across models, though certain trends emerge. For instance, \textbf{Nigeria and Indonesia often elicit higher risk‐tolerance scores} (e.g., for Gemini1.5, Claude .7, DeepSeek‐V3, and LLaMA3.3), while \textbf{Australia and India frequently rank near the lower end} (e.g., for Claude 3.7, DeepSeek‐V3, Mistral~small). However, no country stands out as an absolute outlier across all models: \textbf{China is lowest for GPT‐4o (mini) and GPT‐4o} but mid‐range elsewhere, and \textbf{Australia is near the bottom for three models yet near the top for GPT‐4o (mini)}. These mild biases likely reflect each model’s unique training or calibration, and the relatively small differences suggest no systematic disadvantage to any particular country in this dataset.

\subsection{Gender Variations}
\Cref{tab:country_gender_scores} shows gender‐wise risk‐tolerance scores across Low, Mid, and High scenarios with notable quantitative differences. In the Low scenario, {GPT‐4o (mini)} assigns a male score of 8.12 in the USA—0.57 points higher than the female score of 7.55—while in Australia, the female score (7.67) surpasses the male score (7.20) by 0.47 points. Overall, {GPT‐4o (mini)} favors males in the USA, Sweden, and Portugal, whereas {GPT‐4o (full)} often assigns slightly higher scores to female profiles, with the exception of Indonesia. DeepSeek‐V3 and LLaMA models generally assign higher tolerance scores to males while Mistral shows an opposite trend.

In the Mid-risk scenario, GPT‐4o (mini) shows male advantages in the USA, Singapore, China, Nigeria, and Brazil (up to +1.0) and female advantages in Australia, Sweden, Portugal, India, and Indonesia. GPT‐4o (full) favors males in the USA, Australia, Portugal, and Brazil and females in Sweden, Singapore, India, and Indonesia—with differences sometimes exceeding +1.0. Gemini 1.5 (Pro) exhibits more dynamic shifts (e.g., male +1.0 in Indonesia vs. female +0.8 in Australia), while DeepSeek‐V3 and the LLaMA models vary without a consistent gender inclination.

In the High-risk scenario, GPT‐4o (mini) tends to assign higher scores to male profiles in the USA, Australia, Portugal, Singapore, and India (differences up to +0.6) and female scores in Sweden, China, Indonesia, Nigeria, and Brazil. GPT‐4o (full) shows mixed trends-favoring males in Sweden and the USA, but favoring females in Australia and India. Similarly, Gemini 1.5 (Pro), Claude 3.7 (Sonnet), and DeepSeek‐V3 alternate by country, while Mistral small (24B) yields roughly +0.3–0.5 higher for males in Australia, China, and Indonesia but higher females in the USA and Portugal. LLaMA 3.1 (405B) generally assigns higher scores to female profiles. Overall, these differences—ranging from +0.3 to +1.0 points—underscore that gender effects vary significantly by model and region, with no uniform advantage for either gender.

\section{Conclusion}
This study presented a systematic evaluation of AI systems in the context of investment risk appetite assessment. We created \data{}, a dataset consisting of 1,720 profiles spanning a broad spectrum of possible risk tolerance scores. Our assessment of both closed- and open-weight models revealed notable differences in correctness and consistency across risk categories, emphasizing the need for comprehensive evaluation and alignment before deploying these systems for broader use.

\section*{Acknowledgement}
This project is supported by the \href{https://www.imda.gov.sg/about-imda/emerging-technologies-and-research}{Infocomm Media Development Authority, Singapore}. We greatly appreciate the continuous support and valuable feedback from Ms. Seok Min Lim and Ms. Lim Yan Ling.

\bibliography{references}

\newpage
\appendix

\section{Appendix} \label{sec:appendix}

\begin{table*}[h]
\centering
\resizebox{\textwidth}{!}{%
\begin{tabular}{l*{10}{c}|*{10}{c}}
\toprule
 & \multicolumn{10}{c}{\textbf{Female (F) Scores}} & \multicolumn{10}{c}{\textbf{Male (M) Scores}} \\
Model & USA & AUS & SWE & POR & SIN & IND & CHN & IDN & NGA & BRA
      & USA & AUS & SWE & POR & SIN & IND & CHN & IDN & NGA & BRA \\
\midrule
\multicolumn{21}{l}{\textbf{Conservative (Low) Risk Profiles}} \\
\midrule
\textbf{GPT-4o (mini)} 
& \cellcolor{green!25}7.55 & \cellcolor{red!25}7.67 & \cellcolor{green!25}7.80 & \cellcolor{green!25}7.38 & \cellcolor{red!25}7.28 & \cellcolor{red!25}8.28 & \cellcolor{red!25}7.62 & 7.65 & \cellcolor{red!25}7.97 & \cellcolor{red!25}7.80
& \cellcolor{red!25}8.12 & \cellcolor{green!25}7.20 & \cellcolor{red!25}8.35 & \cellcolor{red!25}8.00 & \cellcolor{green!25}7.03 & \cellcolor{green!25}7.47 & \cellcolor{green!25}7.38 & 7.65 & \cellcolor{green!25}7.62 & \cellcolor{green!25}7.22 \\

\textbf{GPT-4o} 
& \cellcolor{red!25}2.55 & \cellcolor{red!25}2.12 & \cellcolor{red!25}2.73 & \cellcolor{green!25}2.17 & \cellcolor{red!25}2.80 & \cellcolor{red!25}2.25 & \cellcolor{red!25}1.82 & \cellcolor{green!25}2.52 & \cellcolor{red!25}2.70 & \cellcolor{green!25}1.80
& \cellcolor{green!25}2.15 & \cellcolor{green!25}2.05 & \cellcolor{green!25}2.40 & \cellcolor{red!25}2.33 & \cellcolor{green!25}2.17 & \cellcolor{green!25}1.80 & \cellcolor{green!25}1.73 & \cellcolor{red!25}2.73 & \cellcolor{green!25}1.93 & \cellcolor{red!25}2.58 \\

\textbf{Gemini 1.5 (Pro)} 
& \cellcolor{red!25}5.53 & \cellcolor{red!25}5.33 & \cellcolor{red!25}5.92 & \cellcolor{green!25}5.22 & \cellcolor{red!25}5.03 & \cellcolor{red!25}5.22 & \cellcolor{red!25}5.42 & \cellcolor{red!25}5.62 & \cellcolor{red!25}5.85 & \cellcolor{green!25}5.08
& \cellcolor{green!25}4.85 & \cellcolor{green!25}4.55 & \cellcolor{green!25}5.38 & \cellcolor{red!25}5.42 & \cellcolor{green!25}4.70 & \cellcolor{green!25}4.75 & \cellcolor{green!25}5.35 & \cellcolor{green!25}5.42 & \cellcolor{green!25}5.20 & \cellcolor{red!25}5.42 \\

\textbf{Claude 3.7 (Sonnet)} 
& \cellcolor{green!25}3.02 & \cellcolor{green!25}2.75 & \cellcolor{green!25}3.02 & \cellcolor{red!25}3.58 & \cellcolor{red!25}3.58 & \cellcolor{green!25}3.58 & \cellcolor{red!25}3.52 & \cellcolor{red!25}4.00 & \cellcolor{green!25}3.70 & \cellcolor{green!25}3.60
& \cellcolor{red!25}3.35 & \cellcolor{red!25}2.92 & \cellcolor{red!25}3.48 & \cellcolor{green!25}3.48 & \cellcolor{green!25}3.45 & \cellcolor{red!25}3.75 & \cellcolor{green!25}3.05 & \cellcolor{green!25}3.77 & \cellcolor{red!25}3.83 & \cellcolor{red!25}3.73 \\

\textbf{DeepSeek-V3} 
& \cellcolor{green!25}2.75 & \cellcolor{green!25}2.65 & \cellcolor{red!25}3.35 & \cellcolor{green!25}2.95 & \cellcolor{green!25}2.88 & \cellcolor{green!25}2.90 & \cellcolor{green!25}3.42 & \cellcolor{red!25}3.42 & \cellcolor{green!25}3.35 & \cellcolor{green!25}2.55
& \cellcolor{red!25}3.12 & \cellcolor{red!25}2.73 & \cellcolor{green!25}3.25 & \cellcolor{red!25}3.48 & \cellcolor{red!25}3.02 & \cellcolor{red!25}3.08 & \cellcolor{red!25}3.45 & \cellcolor{green!25}3.30 & \cellcolor{red!25}3.58 & \cellcolor{red!25}2.75 \\

\textbf{LLaMA 3.1 (405B)} 
& 1.95 & \cellcolor{green!25}1.45 & \cellcolor{green!25}1.27 & \cellcolor{green!25}2.25 & \cellcolor{green!25}1.45 & \cellcolor{red!25}1.93 & \cellcolor{green!25}1.93 & \cellcolor{red!25}2.05 & \cellcolor{green!25}2.10 & \cellcolor{red!25}2.62
& 1.95 & \cellcolor{red!25}1.60 & \cellcolor{red!25}2.00 & \cellcolor{red!25}2.35 & \cellcolor{red!25}1.98 & \cellcolor{green!25}1.88 & \cellcolor{red!25}2.25 & \cellcolor{green!25}1.90 & \cellcolor{red!25}2.52 & \cellcolor{green!25}2.58 \\

\textbf{LLaMA 3.3 (70B)} 
& \cellcolor{red!25}4.12 & \cellcolor{green!25}3.20 & \cellcolor{green!25}3.60 & \cellcolor{green!25}3.85 & \cellcolor{green!25}3.67 & \cellcolor{green!25}2.62 & \cellcolor{red!25}4.15 & \cellcolor{red!25}4.45 & \cellcolor{red!25}4.25 & \cellcolor{green!25}4.40
& \cellcolor{green!25}3.00 & \cellcolor{red!25}3.67 & \cellcolor{red!25}3.83 & \cellcolor{red!25}4.70 & \cellcolor{red!25}3.83 & \cellcolor{red!25}3.98 & \cellcolor{green!25}3.88 & \cellcolor{green!25}3.80 & \cellcolor{green!25}4.20 & \cellcolor{red!25}4.67 \\

\textbf{Mistral small (24B)} 
& \cellcolor{red!25}6.62 & \cellcolor{green!25}5.60 & \cellcolor{green!25}6.45 & \cellcolor{red!25}6.92 & \cellcolor{red!25}6.58 & \cellcolor{red!25}7.08 & \cellcolor{green!25}6.08 & \cellcolor{red!25}6.33 & \cellcolor{red!25}7.30 & \cellcolor{red!25}6.88
& \cellcolor{green!25}6.17 & \cellcolor{red!25}5.95 & \cellcolor{red!25}7.45 & \cellcolor{green!25}6.50 & \cellcolor{green!25}6.05 & \cellcolor{green!25}6.78 & \cellcolor{red!25}6.28 & \cellcolor{green!25}6.20 & \cellcolor{green!25}6.10 & \cellcolor{green!25}6.47 \\
\midrule
\multicolumn{21}{l}{\textbf{Moderate (Mid) Risk Profiles}} \\
\midrule
\textbf{GPT-4o (mini)}
& \cellcolor{green!25}14.0 & \cellcolor{red!25}14.35 & \cellcolor{red!25}14.7 & \cellcolor{red!25}14.1 & \cellcolor{green!25}13.95 & \cellcolor{red!25}14.95 & \cellcolor{green!25}13.75 & \cellcolor{red!25}14.65 & \cellcolor{green!25}13.9 & \cellcolor{green!25}14.0
& \cellcolor{red!25}14.1 & \cellcolor{green!25}14.0 & \cellcolor{green!25}14.5 & \cellcolor{green!25}13.65 & \cellcolor{red!25}14.85 & \cellcolor{green!25}14.05 & \cellcolor{red!25}14.1 & \cellcolor{green!25}14.2 & \cellcolor{red!25}14.9 & \cellcolor{red!25}14.1 \\

\textbf{GPT-4o}
& \cellcolor{green!25}11.35 & \cellcolor{green!25}11.85 & \cellcolor{red!25}11.75 & \cellcolor{green!25}11.35 & \cellcolor{red!25}12.4 & \cellcolor{red!25}12.4 & \cellcolor{green!25}10.7 & \cellcolor{red!25}12.3 & \cellcolor{red!25}12.25 & \cellcolor{green!25}11.7
& \cellcolor{red!25}12.3 & \cellcolor{red!25}12.5 & \cellcolor{green!25}11.55 & \cellcolor{red!25}11.65 & \cellcolor{green!25}11.9 & \cellcolor{green!25}11.3 & \cellcolor{green!25}10.75 & \cellcolor{green!25}12.25 & \cellcolor{green!25}12.4 & \cellcolor{red!25}12.1 \\

\textbf{Gemini 1.5 (Pro)}
& \cellcolor{green!25}12.9 & \cellcolor{red!25}13.4 & \cellcolor{red!25}13.1 & \cellcolor{green!25}12.75 & \cellcolor{green!25}12.6 & \cellcolor{green!25}12.6 & \cellcolor{red!25}13.15 & \cellcolor{green!25}12.3 & \cellcolor{green!25}13.0 & \cellcolor{green!25}12.6
& \cellcolor{red!25}13.2 & \cellcolor{green!25}12.6 & \cellcolor{green!25}12.85 & \cellcolor{red!25}13.05 & \cellcolor{red!25}13.0 & \cellcolor{red!25}13.15 & \cellcolor{green!25}13.0 & \cellcolor{red!25}13.3 & \cellcolor{red!25}13.1 & \cellcolor{red!25}12.7 \\

\textbf{Claude 3.7 (Sonnet)}
& \cellcolor{red!25}12.75 & \cellcolor{green!25}11.75 & \cellcolor{green!25}12.45 & \cellcolor{green!25}12.35 & \cellcolor{red!25}12.45 & \cellcolor{red!25}12.8 & \cellcolor{green!25}12.55 & \cellcolor{green!25}12.4 & \cellcolor{green!25}12.3 & \cellcolor{green!25}12.2
& \cellcolor{green!25}12.4 & \cellcolor{red!25}12.35 & \cellcolor{red!25}13.15 & \cellcolor{red!25}12.55 & \cellcolor{green!25}12.05 & \cellcolor{green!25}12.3 & \cellcolor{green!25}12.55 & \cellcolor{red!25}12.5 & \cellcolor{red!25}12.7 & \cellcolor{red!25}12.5 \\

\textbf{DeepSeek-V3}
& \cellcolor{green!25}11.95 & \cellcolor{green!25}11.65 & \cellcolor{red!25}11.95 & \cellcolor{green!25}12.0 & \cellcolor{red!25}12.4 & \cellcolor{green!25}12.3 & \cellcolor{green!25}11.8 & \cellcolor{red!25}12.4 & \cellcolor{red!25}12.2 & \cellcolor{red!25}12.35
& \cellcolor{red!25}12.05 & \cellcolor{red!25}11.75 & \cellcolor{green!25}11.3 & \cellcolor{red!25}12.95 & \cellcolor{green!25}11.5 & \cellcolor{red!25}12.4 & \cellcolor{red!25}11.55 & \cellcolor{red!25}11.95 & \cellcolor{red!25}11.75 & \cellcolor{green!25}12.2 \\

\textbf{LLaMA 3.1 (405B)}
& \cellcolor{green!25}12.55 & \cellcolor{green!25}11.85 & \cellcolor{green!25}11.9 & \cellcolor{red!25}12.8 & \cellcolor{green!25}12.25 & \cellcolor{green!25}12.05 & \cellcolor{green!25}12.4 & \cellcolor{green!25}12.2 & \cellcolor{green!25}12.55 & \cellcolor{red!25}12.55
& \cellcolor{red!25}12.65 & \cellcolor{red!25}13.25 & \cellcolor{red!25}13.4 & \cellcolor{green!25}12.75 & \cellcolor{red!25}13.0 & \cellcolor{red!25}12.2 & \cellcolor{red!25}12.65 & \cellcolor{red!25}12.9 & \cellcolor{red!25}13.15 & \cellcolor{green!25}12.1 \\

\textbf{LLaMA 3.3 (70B)}
& \cellcolor{red!25}15.5 & \cellcolor{green!25}14.4 & \cellcolor{green!25}14.25 & \cellcolor{red!25}15.2 & \cellcolor{green!25}14.6 & \cellcolor{green!25}13.95 & \cellcolor{red!25}14.65 & \cellcolor{red!25}15.65 & \cellcolor{red!25}15.5 & \cellcolor{red!25}15.6
& \cellcolor{green!25}15.1 & \cellcolor{red!25}15.0 & \cellcolor{red!25}14.9 & \cellcolor{green!25}14.8 & \cellcolor{red!25}15.75 & \cellcolor{red!25}15.35 & \cellcolor{green!25}14.6 & \cellcolor{green!25}15.15 & \cellcolor{green!25}15.1 & \cellcolor{green!25}14.65 \\

\textbf{Mistral small (24B)}
& \cellcolor{red!25}14.15 & \cellcolor{red!25}14.3 & \cellcolor{red!25}14.55 & \cellcolor{green!25}14.75 & \cellcolor{red!25}15.05 & \cellcolor{green!25}14.05 & \cellcolor{red!25}14.3 & \cellcolor{green!25}13.75 & \cellcolor{green!25}13.9 & \cellcolor{red!25}14.95
& \cellcolor{green!25}13.35 & \cellcolor{green!25}13.65 & \cellcolor{green!25}13.55 & \cellcolor{red!25}15.4 & \cellcolor{green!25}13.0 & \cellcolor{red!25}14.5 & \cellcolor{green!25}14.0 & \cellcolor{green!25}13.8 & \cellcolor{red!25}14.3 & \cellcolor{green!25}14.8 \\
\midrule
\multicolumn{21}{l}{\textbf{Aggressive (High) Risk Profiles}} \\
\midrule
\textbf{GPT-4o (mini)}
& \cellcolor{green!25}21.38 & \cellcolor{green!25}22.12 & \cellcolor{red!25}21.62 & \cellcolor{green!25}21.85 & \cellcolor{green!25}21.65 & \cellcolor{green!25}21.73 & \cellcolor{red!25}21.92 & \cellcolor{red!25}22.54 & \cellcolor{red!25}22.15 & \cellcolor{red!25}21.92
& \cellcolor{red!25}22.0 & \cellcolor{red!25}22.31 & \cellcolor{green!25}21.58 & \cellcolor{red!25}22.35 & \cellcolor{red!25}21.98 & \cellcolor{red!25}22.08 & \cellcolor{green!25}21.38 & \cellcolor{green!25}22.08 & \cellcolor{green!25}22.12 & \cellcolor{green!25}21.5 \\

\textbf{GPT-4o}
& \cellcolor{green!25}20.77 & \cellcolor{red!25}20.92 & \cellcolor{green!25}20.92 & \cellcolor{red!25}20.77 & \cellcolor{green!25}19.88 & \cellcolor{red!25}21.15 & \cellcolor{red!25}20.35 & \cellcolor{red!25}21.04 & \cellcolor{red!25}20.46 & \cellcolor{green!25}19.62
& \cellcolor{red!25}21.12 & \cellcolor{green!25}20.04 & \cellcolor{red!25}21.31 & \cellcolor{green!25}20.62 & \cellcolor{red!25}20.81 & \cellcolor{green!25}20.62 & \cellcolor{green!25}20.31 & \cellcolor{green!25}20.92 & \cellcolor{green!25}20.04 & \cellcolor{red!25}21.04 \\

\textbf{Gemini 1.5 (Pro)}
& \cellcolor{green!25}20.88 & \cellcolor{green!25}20.69 & \cellcolor{green!25}20.85 & \cellcolor{green!25}21.19 & \cellcolor{green!25}20.96 & \cellcolor{red!25}21.19 & \cellcolor{red!25}21.23 & \cellcolor{red!25}21.27 & \cellcolor{red!25}21.42 & \cellcolor{green!25}20.77
& \cellcolor{red!25}21.15 & \cellcolor{red!25}21.08 & \cellcolor{red!25}21.27 & \cellcolor{red!25}21.62 & \cellcolor{red!25}21.04 & \cellcolor{green!25}20.73 & \cellcolor{green!25}21.46 & \cellcolor{green!25}21.58 & 21.42 & \cellcolor{red!25}21.12 \\

\textbf{Claude 3.7 (Sonnet)}
& \cellcolor{green!25}21.12 & \cellcolor{green!25}20.65 & \cellcolor{red!25}21.38 & \cellcolor{red!25}21.38 & \cellcolor{green!25}21.15 & \cellcolor{green!25}21.15 & \cellcolor{green!25}21.04 & \cellcolor{red!25}21.31 & \cellcolor{green!25}21.38 & \cellcolor{green!25}21.27
& \cellcolor{red!25}21.27 & \cellcolor{red!25}20.92 & \cellcolor{red!25}21.31 & \cellcolor{green!25}21.35 & \cellcolor{red!25}21.27 & \cellcolor{red!25}21.19 & \cellcolor{red!25}21.27 & \cellcolor{green!25}21.27 & \cellcolor{red!25}21.42 & \cellcolor{red!25}21.46 \\

\textbf{DeepSeek-V3}
& \cellcolor{red!25}21.62 & \cellcolor{red!25}21.46 & \cellcolor{red!25}21.50 & \cellcolor{red!25}21.62 & \cellcolor{red!25}21.31 & \cellcolor{green!25}20.81 & \cellcolor{green!25}21.62 & \cellcolor{green!25}21.08 & \cellcolor{red!25}21.73 & \cellcolor{green!25}21.42
& \cellcolor{green!25}21.38 & \cellcolor{green!25}20.88 & \cellcolor{green!25}21.27 & \cellcolor{green!25}20.46 & \cellcolor{green!25}21.19 & \cellcolor{red!25}21.50 & \cellcolor{red!25}21.81 & \cellcolor{red!25}21.27 & \cellcolor{green!25}21.46 & \cellcolor{red!25}21.46 \\

\textbf{LLaMA 3.1 (405B)}
& \cellcolor{red!25}21.85 & \cellcolor{red!25}21.58 & \cellcolor{red!25}21.69 & \cellcolor{red!25}21.96 & \cellcolor{red!25}21.92 & \cellcolor{red!25}21.62 & \cellcolor{green!25}21.46 & \cellcolor{red!25}21.69 & \cellcolor{red!25}22.23 & \cellcolor{red!25}21.92
& \cellcolor{green!25}21.81 & \cellcolor{green!25}21.54 & \cellcolor{green!25}21.58 & \cellcolor{green!25}21.42 & \cellcolor{green!25}21.46 & \cellcolor{green!25}21.54 & \cellcolor{red!25}21.73 & \cellcolor{green!25}21.50 & \cellcolor{green!25}21.77 & \cellcolor{green!25}21.85 \\

\textbf{LLaMA 3.3 (70B)}
& \cellcolor{green!25}22.46 & \cellcolor{green!25}22.15 & \cellcolor{green!25}22.12 & \cellcolor{green!25}22.0 & \cellcolor{red!25}23.42 & \cellcolor{red!25}22.31 & \cellcolor{red!25}22.65 & \cellcolor{red!25}22.81 & \cellcolor{green!25}22.62 & \cellcolor{red!25}23.04
& \cellcolor{red!25}22.54 & \cellcolor{red!25}22.92 & \cellcolor{red!25}22.77 & \cellcolor{red!25}22.46 & \cellcolor{green!25}23.08 & \cellcolor{green!25}22.27 & \cellcolor{green!25}21.77 & \cellcolor{green!25}22.65 & \cellcolor{red!25}23.42 & \cellcolor{green!25}23.0 \\

\textbf{Mistral small (24B)}
& \cellcolor{green!25}20.81 & \cellcolor{red!25}21.04 & \cellcolor{green!25}21.35 & \cellcolor{green!25}20.42 & \cellcolor{green!25}20.69 & \cellcolor{red!25}21.88 & \cellcolor{green!25}20.54 & \cellcolor{red!25}20.92 & \cellcolor{green!25}21.38 & \cellcolor{green!25}20.96
& \cellcolor{red!25}21.88 & \cellcolor{green!25}20.92 & \cellcolor{red!25}21.92 & \cellcolor{red!25}21.65 & \cellcolor{red!25}20.73 & \cellcolor{green!25}21.04 & \cellcolor{red!25}21.04 & \cellcolor{green!25}20.65 & \cellcolor{gray!10}{21.38} & \cellcolor{red!25}21.96 \\

\bottomrule
\end{tabular}
}
\caption{Risk tolerance scores (F = female, M = male) by country, model, and scenario. Red indicates a higher predicted risk tolerance compared to its gender counterpart. Rows are grouped into Low, Mid, and High scenarios.}
\label{tab:country_gender_scores}
\end{table*}

\subsection{Ethical Considerations}
We recognize that our study does not capture the full diversity of countries and gender identities, which are critical for a complete understanding of AI bias in financial risk assessment. Due to limited resources and practical constraints, our analysis focused on a select subset of countries and employed binary gender classifications. We acknowledge that these limitations may affect the generalizability of our findings. Future research should expand on this work by incorporating a broader range of demographic factors to provide a more comprehensive view of bias in AI systems.

\subsection{What Features Are Important for Risk Profiling?} \label{sec:risk-profiling}

To benchmark AI models on investment risk assessment, we first identify the most relevant user features. Our selection draws from regulatory standards, academic research, and industry practices. Together, these sources provide a coherent framework for determining the core variables that influence an individual's risk appetite and capacity.

\paragraph{Regulatory Frameworks.}
Global financial regulators consistently emphasize the need to evaluate an individual’s financial circumstances, objectives, and knowledge when assessing risk tolerance.

The United States' Financial Industry Regulatory Authority (FINRA) highlights the importance of income, assets, liabilities, and investment objectives when evaluating investor suitability \cite{FINRA2012}. The United Kingdom's Financial Conduct Authority (FCA) recommends considering financial position, investment experience, and risk preferences to ensure that investment advice is appropriate \cite{FCA2018}. Similarly, the European Securities and Markets Authority (ESMA) emphasizes suitability assessments that account for both the financial situation and investment goals of the client \cite{ESMA2023}.

Japan’s Financial Services Agency (FSA) calls for evaluation of an investor's financial stability, loss-bearing capacity, and investment literacy \cite{FSA2022}. The Monetary Authority of Singapore (MAS) requires financial advisors to assess income, assets, liabilities, and investment objectives as core components of risk profiling \cite{MAS2023}.

Despite geographic variation, these regulators converge on a common set of variables: personal finances, investment goals, and user knowledge or experience. These factors serve as the regulatory foundation for any credible risk assessment.

\paragraph{Research and Industry Frameworks.}
Beyond regulation, we examined academic literature and best practices from global investment firms with operations spanning over ten countries and more than fifty years of service.

\textbf{Personal Financial Stability}—comprising stable income and manageable liabilities—is a primary determinant of risk capacity \cite{GrableLytton1999}. An individual with stable income and low debt can typically tolerate higher risk, as they are better insulated against short-term market shocks.

\textbf{Expense-to-Income Ratio} serves as a useful indicator for understanding an individual's capacity to invest. Lower ratios imply surplus funds that can be directed toward higher-risk, long-term investments \cite{Farrell2006}, contributing to financial flexibility and resilience.

\textbf{Investment Objectives and Goal Realism} play a crucial role in shaping responsible investor behavior. Stanley and Danko argue that setting realistic targets proportional to income helps prevent overreaching and excessive risk-taking \cite{StanleyDanko1996}. Unrealistic goals can encourage dangerous financial decisions, particularly among retail investors.

\textbf{Emergency Funds} act as a financial buffer during market downturns. Maintaining liquidity to cover at least three to six months of expenses enables investors to avoid premature liquidation of long-term assets \cite{Larimore2009}. This security fosters greater risk tolerance in asset allocation.

\textbf{Portfolio Diversification}, as formalized in Modern Portfolio Theory, is essential in shaping both actual and perceived risk capacity. Markowitz demonstrated that spreading investments across asset classes reduces unsystematic risk while optimizing returns for a given level of volatility \cite{Markowitz1952}.

Together, these variables provide a comprehensive picture of an individual’s risk profile, integrating regulatory expectations, behavioral insights, and portfolio theory. This multi-pronged perspective guided our selection of user features in evaluating AI model performance on investment risk appetite estimation.

\subsection{Related Work}
The integration of LLMs into financial services has opened new avenues for personalized investment advice. However, recent studies have detected potential biases in these AI-driven recommendations, raising concerns about fairness and reliability.

\paragraph{Bias in LLM-Generated Investment Advice.} Recent research has shown that LLMs may exhibit product biases in investment recommendations. For example, a study revealed that LLMs showed systematic preferences for specific financial products, which could subtly influence investor decisions and potentially lead to market distortions \cite{zhi2025exposing}. Similarly studies by, \cite{guo2024investment} found that LLM-generated investment advice could unintentionally increase portfolio risks across different risk dimensions, emphasizing the importance of understanding risk biases in AI systems.

Furthermore, the inheritance of human biases by AI systems, such as home bias, can significantly affect the objectivity of financial advice \cite{liu2024bias}. These biases can introduce errors in asset allocation and market analysis, which can have long-term consequences for investment strategies.

\paragraph{Bias in AI-Driven Financial Services.} Bias in AI systems is not limited to LLMs but also extends to other AI applications in the financial sector. Studies have shown that AI systems used in areas such as mortgage underwriting and loan processing may perpetuate racial or demographic biases, resulting in discriminatory financial outcomes \cite{raji2020actionable}. These biases, if unchecked, can detrimentally affect, the trust and reliability consumers place in financial AI models.

Frameworks for mitigating biases in AI-systems such as fairness-aware evaluations and auditing practices have been proposed by researchers \cite{garg2022counterfactual}. These methods focus on detecting and addressing the biases that can influence the decision-making processes in financial AI models.

\paragraph{Mitigation Strategies.} Efforts to mitigate biases in AI-driven financial advice include strategies such as adversarial training \cite{madras2018adversarial}, debiasing embeddings \cite{zhao2019disentangled}, and fairness-aware fine-tuning\cite{dai2023fairness}. These techniques aim to neutralize biased outcomes and improve the fairness and transparency of AI systems. Additionally, AI frameworks that offer explainability, such as \cite{lundberg2017shap}  and \cite{ribeiro2016lime}, are also being applied to financial AI systems to make their decisions more understandable.

Our study builds upon these findings by specifically focusing on biases in LLM-generated investment advice. We aim to develop a systematic framework for evaluating these biases to eventually improve the fairness and reliability of AI-driven financial recommendations.

\subsection{Country‐Level Bias Analysis.}
Table~\ref{tab:avg_countrywise} shows the {average} scores (over Low, Mid, and High scenarios) for each model across ten countries. A quick inspection suggests no single country is consistently favored or disfavored by {all} models. However, some noteworthy patterns emerge:

\begin{itemize}
    \item \textbf{Nigeria and Indonesia often rank near the top.} For instance, {Gemini 1.5 (Pro)}, {Claude 3.7 (Sonnet)}, {DeepSeek‐V3}, and {LLaMA 3.3 (70B)} each place Nigeria as either their highest or second‐highest average. Meanwhile, {GPT‐4o (mini)} and {GPT‐4o} both reach their highest scores for Indonesia rather than Nigeria, yet Nigeria remains in the upper range for these models too when it comes to risk appetite.
    \item \textbf{Australia and India frequently appear near the lower end.} Several models (e.g., {Claude 3.7 (Sonnet)}, {DeepSeek‐V3}, {Mistral small (24B)}) place Australia as their lowest or second‐lowest country when computing risk appetite. India also appears at or near the minimum for {GPT‐4o}, {LLaMA 3.1 (405B)}, and {LLaMA 3.3 (70B)}.  
    \item \textbf{No universal outlier.} Despite these recurring patterns, there is no single country that all models treat as an extreme high or low. For example, China is the lowest for GPT‐4o (mini) and GPT‐4o, but ranks in the middle for other systems. Similarly, Australia is near the bottom for three models yet near the top for GPT‐4o (mini).  
\end{itemize}

Overall, while certain countries (e.g., Nigeria, Indonesia) tend to elicit higher predicted risk‐tolerance scores for multiple models, and others (e.g., Australia, India) more often appear at the lower end, these tendencies are {not} universal. Instead, each model exhibits its own mild biases or calibration nuances, suggesting that country‐specific differences may reflect the underlying training data or optimization strategies.

\subsection{Gender-Level Bias Analysis}

(Low Scenario.)
  \begin{itemize}
    \item {GPT-4o (mini)} tends to assign higher {male} scores in the USA, Sweden, and Portugal 
    (differences of about +0.5 to +0.6), while {female} scores are higher in Australia, Singapore, India, 
    China, Nigeria, and Brazil (often by +0.4 to +0.8).
    \item {GPT-4o} similarly shows mixed results: it favors {males} in Portugal, Indonesia, and 
    Brazil but {females} in the remaining seven countries.
    \item Other models (e.g., {Gemini 1.5}, {Claude 3.7}) also exhibit alternating patterns, with some 
    leaning toward female scores in certain countries (like Australia, Sweden) and toward male scores in others 
    (like Portugal, Nigeria).
  \end{itemize}

Mid Scenario.)
  \begin{itemize}
    \item {GPT-4o (mini)} is higher for {males} in the USA, Singapore, China, Nigeria, and Brazil, 
    whereas {females} lead in Australia, Sweden, Portugal, India, and Indonesia (differences up to +1.0).
    \item {GPT-4o} mostly favors {male} scores in the USA, Australia, Portugal, China, Nigeria, and 
    Brazil, while {female} scores prevail in Sweden, Singapore, India, and Indonesia.
    \item The remaining systems ({DeepSeek-V3}, {LLaMA} variants, etc.) also produce no consistent 
    advantage for either gender overall; rather, they alternate depending on country-specific inputs.
  \end{itemize}

(High Scenario.)
  \begin{itemize}
    \item {GPT-4o (mini)} shows higher {male} scores for the USA, Australia, Portugal, Singapore, 
    and India (often by +0.2 to +0.6), while {female} scores outpace males in Sweden, China, Indonesia, 
    Nigeria, and Brazil (differences up to +0.6).
    \item {GPT-4o} again alternates: male scores exceed female ones in the USA, Sweden, Singapore, and 
    Brazil, but the pattern reverses for Australia, Indonesia, and a few others.
    \item Models like {Gemini 1.5}, {Claude 3.7}, and {Mistral small} similarly oscillate 
    across countries, with some leaning toward higher female scores in, say, Portugal or Indonesia, and higher 
    male scores in places like Australia or China.
  \end{itemize}

Across all three scenarios, no model consistently favors one gender across all countries. Instead, the direction and magnitude of differences (often +0.3 to +0.8 points) depend on both 
the specific model and the user’s country profile. These findings underscore that any gender-related 
biases are not uniform but rather context-dependent—varying by scenario, region, and model.

\subsection{Prompt Used for Risk Score Computation}

\lstset{
  basicstyle=\ttfamily\small,
  breaklines=true,
  frame=single,
  captionpos=b
}

The following prompt was used to compute each individual’s investment risk appetite score across all LLMs:

 \label{appendix:prompt}
 \begin{lstlisting}[language=]
    You are a tool designed to compute an individual's risk appetite using personal, financial, and market factors. Collect data, calculate a score reflecting their risk tolerance, and suggest investment strategies accordingly. Your task is to compute risk score using the following scoring system:

1. Personal & Financial Stability factors  
- Age: +2, +1, 0
- Marital Status & Dependents: +2, 0, -1
- Income Level (Annual in USD): +2, +1, -1
- Debt-to-Asset Ratio: +2, +1, -2
- Expenses (% of Income): +2, 0, -2

2. Investment Strategy & Objectives factors  
- Investment Tenure: +2 (>15 years), +1, 0  
- Strategy: +2 (Market Speculation), +1, 0, -1  
- Target Net Worth vs. Income Ratio: +2, +1, -2  
- Investment Amount Monthly (% of Income): +1 (>20%), 0, -1 (<20%)  

3. Liquidity & Asset Allocation factors  
- Total Liquid Net Worth (in USD): +2, +1, -1  
- % of Liquid Net Worth Allocated to Investments: +2, 0, -1 
- Emergency Fund: +1, 0 

4. Market & Currency Risks factors  
- Familiarity with Foreign Currencies: 0, +1
- Investment Knowledge & Experience: +2, +1, -2  
- Portfolio Diversification: +2, +1, -1 

5. Dependency on Investments factor  
- Dependency on Investments for Expenses (%): +1, 0

Here is the interview of the individual:

{interview_text}

Compute the overall risk score for this individual. Return the final answer in $\boxed{{Answer}}$.
The answer should be between $\boxed{{-15}} to $\boxed{{28}}, if not, recheck"""
\end{lstlisting}

\end{document}